\documentclass[11pt]{article}

\usepackage[margin=1in]{geometry}
\usepackage[T1]{fontenc}
\usepackage{lmodern}
\usepackage{microtype}
\usepackage{amsmath,amssymb}
\usepackage{booktabs}
\usepackage{graphicx}
\usepackage{xcolor}
\usepackage{tikz}
\usetikzlibrary{arrows.meta,positioning}
\usepackage{enumitem}
\usepackage{caption}
\usepackage{float}
\usepackage{fancyhdr}
\usepackage{natbib}
\usepackage[hidelinks]{hyperref}

\definecolor{rrblue}{HTML}{2563EB}
\definecolor{rrgray}{HTML}{4B5563}
\hypersetup{
  colorlinks=true,
  linkcolor=rrblue,
  citecolor=rrblue,
  urlcolor=rrblue,
  pdftitle={Right Reset: Chunking by Prefix Removal},
  pdfauthor={Mike Vegeto},
  pdfsubject={Text chunking by counterfactual prefix removal},
  pdfkeywords={Right Reset, text chunking, causal language models, prefix removal}
}
\setlist{nosep,leftmargin=1.35em}
\newcommand{\RR}{\textsc{Right Reset}}
\newcommand{\KL}{D_{\mathrm{KL}}}
\newcommand{\E}{\mathcal{E}}
\newcommand{\C}{\mathcal{C}}
\newcommand{\method}[1]{\texttt{#1}}

\title{\textbf{Right Reset: Chunking by Prefix Removal}}
\author{Mike Vegeto\\Independent Researcher}
\date{August 2026}

\begin{document}
\maketitle

\begin{abstract}
Removing the left context from a causal language model reveals a useful kind of boundary: an edge where the model processes the same right-hand tokens with little change. We turn this observation into prefix-removal probing and introduce \RR{} (RR), which measures preservation of the right-hand hidden-state trajectory. A dynamic program converts RR edge scores into variable-length chunks. On flattened text formed by concatenating topically similar records after deleting their separators and layout, RR recovers 47.7\% of the original records as clean units, versus 25.9\% for a BGE embedding boundary, the strongest tested conventional baseline without task-specific model training. The gain persists after rendering and OCR. Passive scores from the same Qwen3-4B layer and direct prompting of a same-scale instruction model perform substantially worse on flattened records. Across six language models, RR-selected cuts also undergo consistently less local output disruption than unselected candidate edges. An observed-token likelihood-ratio readout is competitive in some architectures, indicating that the central contribution is the intervention: context dependence itself can provide a boundary signal when surface structure is weak.
\end{abstract}


\section{Introduction}

A language model can process some continuations almost the same way even after everything before them is removed. Those points are natural candidates for chunk boundaries: the right side is locally less dependent on the left.

This signal matters when conventional cues are weak. Most text chunkers rely on paragraphs, sentences, layout, or local semantic change. These are good defaults for ordinary documents, but they can disappear when records are flattened or become ambiguous when neighboring units discuss the same topic. The text may still contain distinct contextual units even when its visible structure no longer reveals them.

\RR{} asks one question at every candidate edge: \emph{if the model started here, how differently would it process what comes next?} It removes the complete left prefix, reprocesses the same right-hand tokens, and compares their hidden-state trajectories with those from uninterrupted processing. A high score means the local trajectory is well preserved. RR therefore finds boundaries through contextual dependence rather than a prescribed document convention.

RR is an edge score, not a chunk-size rule. After scoring, a dynamic program chooses a globally feasible set of cuts. The chunk count may be fixed when comparing score quality or inferred by a cut penalty in an application. This separation lets the same boundary signal support different segmentation policies.

We make three contributions:

\begin{enumerate}
  \item We introduce prefix-removal probing and RR, a hidden-state preservation score that turns local context dependence into a chunking signal.
  \item We show that RR recovers 47.7\% of topically similar source records from flattened text, versus 25.9\% for a BGE embedding boundary and 3.1\% for a same-scale prompted segmenter, and that the result survives rendering and OCR.
  \item We show that the intervention supplies the signal: passive scores from the same model and layer are much weaker, while six-model output analyses confirm that RR selects edges with low local context dependence.
\end{enumerate}

Observed-token likelihood ratio provides a second readout of the same intervention. Its relative strength varies by architecture. This makes the conceptual hierarchy clear: prefix removal is the probing principle, RR is the primary hidden-state readout, dynamic programming turns scores into chunks, and flattened records are the main application test.

\section{Related Work}

Classical text segmentation uses lexical cohesion, discourse structure, or a generative document model. TextTiling detects changes in lexical co-occurrence \citep{hearst1997texttiling}; statistical methods optimize global segmentations \citep{utiyama2001statistical}; and supervised systems learn boundaries from document structure \citep{koshorek2018segmentation}. Applied chunkers use semantic shifts, structured boundaries, or contextual embeddings \citep{duarte2024lumberchunker,li2026boundrl,gunther2024latechunking}. RR adds a different criterion: how much a fixed causal model depends on the prefix at an edge.

Language-model likelihood has also been used for segmentation, including high-loss dialogue units, local perplexity criteria, and context-sensitive tokenization scores \citep{feng2021language,zhao2024metachunking,hronsky2024tokenization}. Segmental models optimize complete-segment likelihoods with dynamic programming \citep{wang2017segmentations}. These methods establish both likelihood and global optimization as useful tools, but they generally score text under its existing context. RR instead deletes the complete prefix and compares aligned processing before and after the intervention.

Context ablation and causal analysis test whether information affects model behavior \citep{vig2020causal,khandelwal2018sharp,oconnor2021context}, while representation analyses show that layer choice changes what hidden states encode \citep{liu2019linguistic,tenney2019bert}. RR turns a specific context ablation into a boundary score and then into a segmentation.

\section{Right Reset}
\label{sec:method}

\subsection{Boundary score}

Let $x_{1:n}$ be a tokenized sequence and let $h^{(\ell)}_j(x_{1:n})\in\mathbb{R}^d$ be the hidden state at position $j$ from layer $\ell$ of a causal language model. Consider the edge $b$ between $x_b$ and $x_{b+1}$. For a right-hand window $W$, define
\[
  r_b=x_{b+1:b+m_b}, \qquad m_b=\min(W,n-b).
\]
We run $r_b$ again after removing $x_{1:b}$, obtaining reset states $\widetilde h^{(\ell,b)}_q$ for $q=0,\ldots,m_b-1$. The default implementation preserves the original position IDs so that token identity and alignment are unchanged; a reindexing control gives equivalent results on Qwen2.5.

RR is the mean cosine preservation of the full and reset trajectories after skipping the first $s$ reset positions:
\begin{equation}
  R_\ell(b)=\frac{1}{m_b-s}\sum_{q=s}^{m_b-1}
  \frac{h^{(\ell)}_{b+1+q}\cdot\widetilde h^{(\ell,b)}_q}
  {\lVert h^{(\ell)}_{b+1+q}\rVert_2\lVert\widetilde h^{(\ell,b)}_q\rVert_2}.
  \label{eq:rr}
\end{equation}
Higher values mean that removing the prefix changes the local hidden trajectory less. The score is directional: it measures dependence of the right side on the left, not symmetric dissimilarity between adjacent spans. Unless stated otherwise, we use $W=24$, $s=1$, and a layer at 75\% of decoder depth.

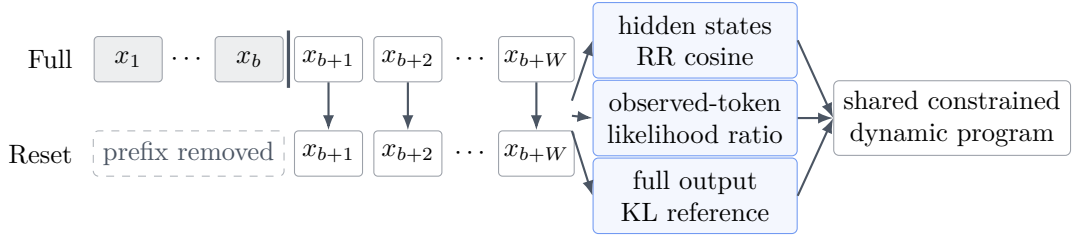
\begin{figure}[t]
\centering
\begin{tikzpicture}[
  font=\small,
  token/.style={draw=rrgray!55, rounded corners=1.5pt, minimum height=6mm, minimum width=9mm, inner sep=2pt},
  readout/.style={draw=rrblue!65, rounded corners=2pt, fill=rrblue!5, align=center, minimum width=27mm, minimum height=10mm},
  flow/.style={-{Latex[length=2mm]}, thick, draw=rrgray}
]
  \node[anchor=east] at (-0.25,0) {Full};
  \node[token,fill=rrgray!10] (l1) at (0.35,0) {$x_1$};
  \node at (1.15,0) {$\cdots$};
  \node[token,fill=rrgray!10] (lb) at (1.95,0) {$x_b$};
  \draw[very thick,rrgray] (2.47,-0.38)--(2.47,0.38);
  \node[token] (r1) at (3.00,0) {$x_{b+1}$};
  \node[token] (r2) at (4.05,0) {$x_{b+2}$};
  \node at (4.90,0) {$\cdots$};
  \node[token] (rw) at (5.75,0) {$x_{b+W}$};

  \node[anchor=east] at (-0.25,-1.25) {Reset};
  \node[draw=rrgray!45,dashed,rounded corners=2pt,minimum width=23mm,minimum height=6mm,align=center,text=rrgray] at (1.15,-1.25) {prefix removed};
  \node[token] (rr1) at (3.00,-1.25) {$x_{b+1}$};
  \node[token] (rr2) at (4.05,-1.25) {$x_{b+2}$};
  \node at (4.90,-1.25) {$\cdots$};
  \node[token] (rrw) at (5.75,-1.25) {$x_{b+W}$};
  \draw[flow] (r1)--(rr1); \draw[flow] (r2)--(rr2); \draw[flow] (rw)--(rrw);

  \node[readout] (rr) at (7.85,0.25) {hidden states\\RR cosine};
  \node[readout] (lr) at (7.85,-0.78) {observed-token\\likelihood ratio};
  \node[readout] (kl) at (7.85,-1.81) {full output\\KL reference};
  \draw[flow] (6.22,-0.55)--(rr.west);
  \draw[flow] (6.22,-0.75)--(lr.west);
  \draw[flow] (6.22,-0.95)--(kl.west);
  \node[draw=rrgray!55,rounded corners=2pt,align=center,minimum width=25mm] (dp) at (11.25,-0.78) {shared constrained\\dynamic program};
  \draw[flow] (rr.east)--(dp.west); \draw[flow] (lr.east)--(dp.west); \draw[flow] (kl.east)--(dp.west);
\end{tikzpicture}
\caption{RR removes the complete prefix but holds the right-hand tokens and their alignment fixed. Hidden-state preservation supplies the RR score. Observed-token likelihood ratio reads the same intervention behaviorally; output KL is used only as a same-model mechanism reference.}
\label{fig:method}
\end{figure}

\subsection{Two other readouts of the intervention}

Let $p^{\mathrm{full}}_{b,q}$ and $p^{\mathrm{reset}}_{b,q}$ be the next-token distributions at aligned full and reset positions. The observed-token likelihood-ratio damage over realized continuation tokens is
\begin{equation}
 d_{\mathrm{LR}}(b)=\frac{1}{W-s-1}\sum_{q=s}^{W-2}
 \log\frac{p^{\mathrm{full}}_{b,q}(x_{b+q+2})}
 {p^{\mathrm{reset}}_{b,q}(x_{b+q+2})}.
 \label{eq:lr}
\end{equation}
We use $-d_{\mathrm{LR}}(b)$ as a boundary utility. This is counterfactual: unlike raw surprisal, it compares the same observed token with and without the prefix. If the next token were sampled from $p^{\mathrm{full}}$, the expectation of the log ratio would equal $\KL(p^{\mathrm{full}}\|p^{\mathrm{reset}})$ \citep{kullback1951information}. Corpus tokens are not such samples, so the observed value is not an unbiased per-position KL estimate.

For evaluation, we measure the complete distributional change:
\begin{equation}
 d_{\mathrm{KL}}(b)=\frac{1}{W-s-1}\sum_{q=s}^{W-2}
 \KL\!\left(p^{\mathrm{full}}_{b,q}\,\|\,p^{\mathrm{reset}}_{b,q}\right).
 \label{eq:kl}
\end{equation}
This KL score is the mechanism reference: zero means that removing the prefix leaves the model's local predictive distribution unchanged.

\subsection{From edge scores to chunks}

RR scores edges independently. We convert each method's valid scores to within-document average-rank percentiles and select cuts with exact dynamic programming. For score-quality comparisons, all methods use the same eligible edges, fixed chunk count $K$, target length $T$, hard limits $L_{\min}$ and $L_{\max}$, and length coefficient $\lambda$. With calibrated utility $u(b)$, the segmentation $0=t_0<\cdots<t_K=n$ maximizes
\begin{equation}
 \sum_{k=1}^{K-1}u(t_k-1)
 -\lambda\sum_{k=1}^{K}\left(\frac{t_k-t_{k-1}-T}{T}\right)^2,
 \label{eq:dp}
\end{equation}
subject to the hard length limits and eligible internal cuts. The bounded recurrence runs in $O(KnL_{\max})$ time with $O(Kn)$ state. Applications may instead leave $K$ open and charge a calibrated penalty for each cut, producing variable chunk sizes and counts.

\section{Flattened-Record Recovery}
\label{sec:application}

When topically similar records are concatenated after their separators and layout are removed, RR recovers 47.7\% of the original records as clean units. A BGE embedding boundary, the strongest tested conventional baseline without task-specific model training, recovers 25.9\%. This is the paper's main application result.

The benchmark contains 60 held-out streams built from 276 unique FiQA, NFCorpus, and SciFact records in BEIR \citep{thakur2021beir}; 15 calibration streams contain 65 additional records. Each stream contains three to six records joined by one ordinary space. BGE-base-en-v1.5 similarity is used to place related records next to one another, weakening topic-shift cues \citep{bgebasecard}. A second construction uses E5-base-v2 similarity, removing BGE's dual role in constructing the benchmark and serving as its strongest conventional baseline \citep{wang2022textembeddings}.

We evaluate flattened original text and a one-column 180-DPI rendering processed by local Apple Vision OCR. Chunkers receive no line, page, bounding-box, whitespace-gap, source-record, or record-count information. All methods use word-boundary candidates on an eight-token grid plus sentence ends, 48--384-token hard limits, and a 192-token soft target. RR uses Qwen3-4B layer 27 and $W=24$. An open-count dynamic program charges a method-specific cut penalty calibrated only on the 15 calibration streams; record counts are not available at inference time.

A predicted chunk is a clean recovered unit when it contains at least 90\% of one source record and at least 90\% of the chunk belongs to that record. Partition F1 is the harmonic mean of token-weighted chunk purity and record completeness. Observed-token LR is the comparison within the prefix-removal family. Conventional baselines without task-specific model training are BGE boundary distance, passive surprisal, sentence cuts, and a fixed grid. Three further controls use the same Qwen layer as RR: residual jump, symmetric local hidden-state distance, and attention isolation.

We also test direct prompting with instruction-tuned Qwen3-4B. It scores twelve tagged candidates at a time using 96 tokens of context on each side and returns a probability for every edge. This gives the prompted model bidirectional evidence while keeping the record count hidden. Its scores use the same candidate set, calibration split, and open-count dynamic program as RR. Statistics use 10,000 domain-stratified paired stream-bootstrap resamples.

\begin{table}[H]
\centering
\scriptsize
\caption{Flattened-record recovery with paired 95\% bootstrap intervals. BGE is the strongest tested conventional baseline without task-specific model training. Prompted Qwen is evaluated on the primary flattened-original construction.}
\label{tab:application}
\resizebox{\textwidth}{!}{%
\begin{tabular}{llccccc}
\toprule
Construction & Text & RR & Observed-token LR & Prompted Qwen & BGE boundary & RR $-$ BGE [95\% CI] \\
\midrule
\multicolumn{7}{l}{\textit{Clean-unit recovery}} \\
BGE-packed & Flattened original & 0.477 & 0.426 & 0.031 & 0.259 & 0.219 [0.125, 0.311] \\
BGE-packed & 180-DPI OCR & 0.487 & 0.404 & --- & 0.269 & 0.218 [0.128, 0.305] \\
E5-packed & Flattened original & 0.490 & 0.417 & --- & 0.306 & 0.183 [0.098, 0.266] \\
\midrule
\multicolumn{7}{l}{\textit{Partition F1}} \\
BGE-packed & Flattened original & 0.893 & 0.878 & 0.730 & 0.823 & 0.070 [0.049, 0.093] \\
BGE-packed & 180-DPI OCR & 0.898 & 0.867 & --- & 0.829 & 0.069 [0.051, 0.087] \\
E5-packed & Flattened original & 0.894 & 0.865 & --- & 0.819 & 0.075 [0.053, 0.097] \\
\bottomrule
\end{tabular}}
\end{table}

The paired RR--BGE interval is positive for both metrics in all three conditions. On flattened original text, the strongest same-Qwen passive score recovers 19.0\% of clean units and reaches 0.797 partition F1, compared with RR's 47.7\% and 0.893. Access to the same representation alone does not explain the gain. The prompted Qwen baseline reaches 3.1\% [1.4\%, 5.2\%] clean-unit recovery and 0.730 [0.717, 0.743] partition F1. RR's paired gains are 44.6 points [37.7, 51.5] and 0.164 [0.146, 0.180], respectively, and are positive within all three domains.

Rendering and OCR leave the RR--BGE advantage essentially unchanged. RR's clean-unit recovery also exceeds LR by 0.0825 [0.0039, 0.1647] in this condition. The E5-packed construction reproduces the result and removes BGE from record selection. OCR is therefore a robustness test of the flattened-text result, not the task definition.

\begin{figure}[t]
\centering
\includegraphics[width=0.76\textwidth]{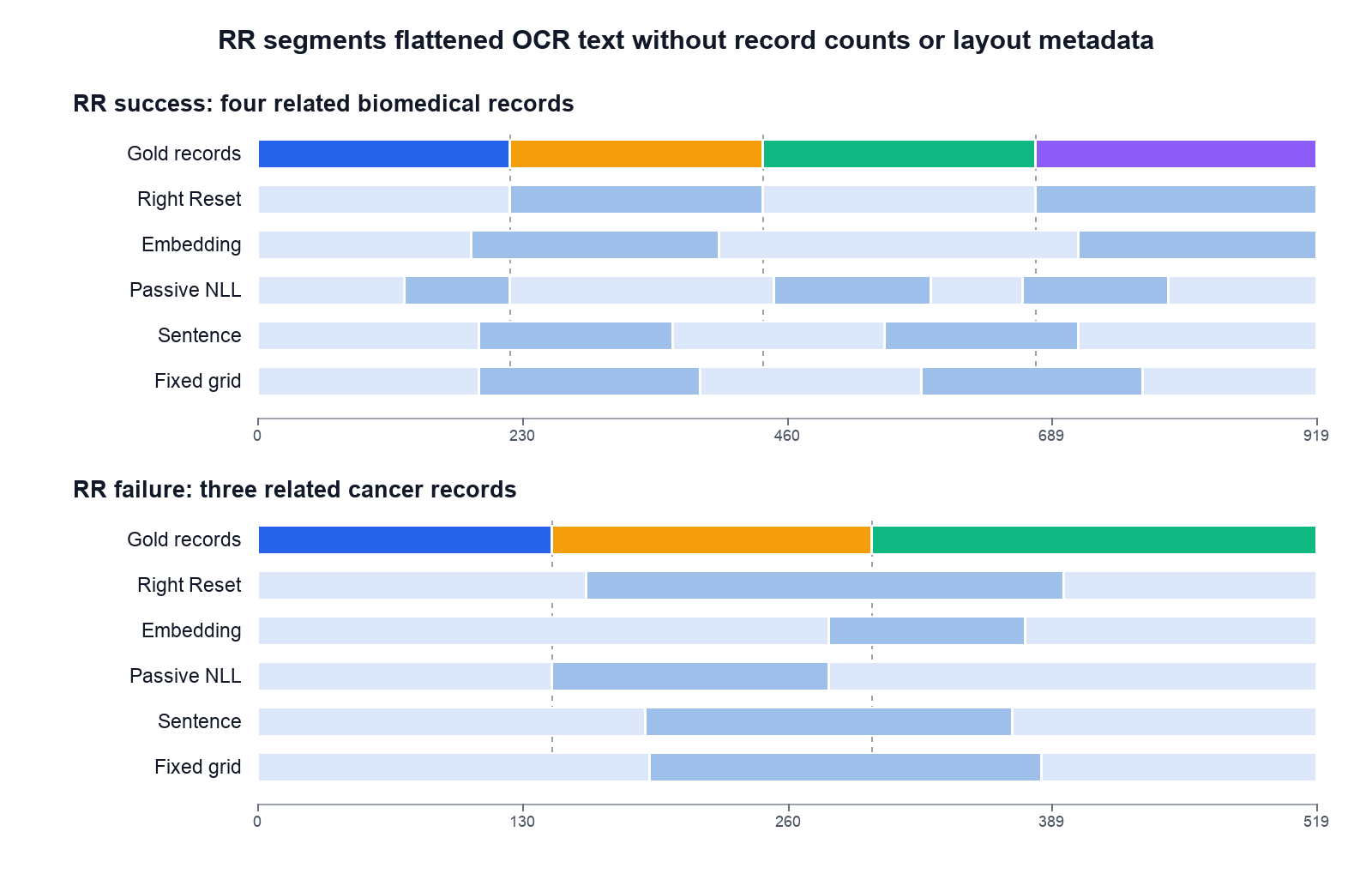}
\caption{Two complete held-out streams. Colored bars are hidden records; pale bars are predicted chunks. RR exactly recovers four related biomedical records in the upper stream. In the lower stream, it misses both joins between three related cancer records while passive surprisal recovers them more closely.}
\label{fig:examples}
\end{figure}

Figure~\ref{fig:examples} shows a complete recovery and a failure, illustrating both the signal and its document-level variation. Exact RR scoring is more expensive than one-pass scoring: on an M3 Max, the Qwen3-4B implementation processes 50,818 source tokens and 7,701 candidates in 289 seconds after warmup, or 176 source tokens per second. On the same hardware, the chunkwise prompted baseline averages 25.0 seconds per flattened-original stream; the RR workload averages 4.8 seconds per OCR stream over the same underlying 60-stream set. This positions the current implementation for offline curation at the tested scale.

As a scope control, we also evaluate conventional top-level sectioning on Wiki-50 \citep{koshorek2018segmentation}. Under development-calibrated thresholds, RR and BGE are statistically indistinguishable ($P_k=0.338$ and $0.340$), and both trail the published supervised Wiki-727K model ($0.182$). This provides no evidence of an RR advantage where sentence and topic cues remain available. Full results and protocol details appear in Appendix~\ref{app:wiki50}.

\section{Evidence for the Boundary Signal}
\label{sec:mechanism}

The application shows that RR finds useful cuts. We next test the proposed mechanism: RR should prefer edges where the model's local predictions depend less on the removed prefix. For selected cuts $\C$ and common eligible edges $\E$, we measure
\begin{equation}
 M_{\mathrm{KL}}(\C)=
 \frac{1}{|\E\setminus\C|}\sum_{b\in\E\setminus\C}d_{\mathrm{KL}}(b)
 -\frac{1}{|\C|}\sum_{b\in\C}d_{\mathrm{KL}}(b).
 \label{eq:margin}
\end{equation}
A positive margin means that selected cuts undergo less output-distribution change than eligible edges left inside chunks. This is a direct test of the prefix-removal mechanism; the flattened-record experiment supplies the external boundary target.

We evaluate two deterministic passages from each of ten literary and technical sources on Qwen3-4B, Qwen3.5-9B, Gemma 4 E4B, Qwen2.5-0.5B, GPT-2, and Pythia-410M \citep{qwen3report,qwen35base,gemma4card,qwen25report,radford2019gpt2,biderman2023pythia}. RR, observed-token LR, and ten non-counterfactual baselines use the same candidates, cut count, length constraints, and dynamic program. Confidence intervals use 10,000 source-cluster bootstrap resamples.

\begin{table}[H]
\centering
\scriptsize
\caption{Output-KL separation margins with 95\% source-cluster bootstrap intervals. ``Passive'' is the strongest non-counterfactual point estimate for that model.}
\label{tab:mechanism}
\resizebox{\textwidth}{!}{%
\begin{tabular}{lccc}
\toprule
Model & RR margin [95\% CI] & Strongest passive & Observed-token LR [95\% CI] \\
\midrule
Qwen3-4B & 0.404 [0.340, 0.482] & 0.094 (attention) & 0.376 [0.282, 0.504] \\
Qwen3.5-9B & 0.442 [0.370, 0.535] & 0.233 (attention) & 0.401 [0.309, 0.512] \\
Gemma 4 E4B & 0.450 [0.335, 0.575] & 0.396 (attention) & 0.655 [0.504, 0.803] \\
Qwen2.5-0.5B & 0.411 [0.337, 0.511] & 0.132 (attention) & 0.352 [0.258, 0.475] \\
GPT-2 & 1.010 [0.873, 1.150] & 0.475 (attention) & 1.475 [1.325, 1.626] \\
Pythia-410M & 0.331 [0.280, 0.396] & 0.082 (window surprisal) & 0.272 [0.210, 0.345] \\
\bottomrule
\end{tabular}}
\end{table}

RR has positive separation in all ten source clusters for each model, and every model-level interval is above zero. It significantly exceeds the strongest passive method in five of six models; the Gemma comparison is inconclusive. RR also significantly exceeds the strongest raw-likelihood baseline in every model.

The observed-token likelihood ratio confirms that the intervention matters beyond hidden-state cosine. RR wins in Qwen2.5 and Pythia, the methods tie in Qwen3 and Qwen3.5, and LR wins in GPT-2 and Gemma. Prefix-removal probing is consistently useful; the strongest readout is architecture-dependent. RR remains attractive because it is effective, reads an intermediate trajectory directly, and avoids vocabulary-wide distribution comparisons.

A layer sweep rises sharply from early layers, peaks near 75\% of decoder depth, and declines at the final layer. The boundary signal therefore develops through the network rather than reducing to an input-level discontinuity. Window and reset-prefix controls appear in Appendix~\ref{app:configuration}.

\section{Discussion}

The experiments support a simple interpretation. Causal models contain boundary information in how their right-hand processing changes when earlier context is removed. Passive hidden-state jumps, attention isolation, raw likelihood, and semantic distance recover much less of that information. RR measures it through hidden-state preservation; observed-token likelihood ratio shows that the signal is also visible behaviorally.

The hidden-state and likelihood readouts have different advantages. RR avoids vocabulary normalization and uses a representation available before the output projection. Observed-token LR has a direct behavioral interpretation and wins in some architectures. Together they show that complete prefix removal is the central operation, with RR as the primary hidden-state readout.

The Wiki-50 scope control reinforces the intended boundary of the claim: contextual dependence is most useful when separators disappear and neighboring units remain topically similar, not as a replacement for conventional chunkers.

\section{Limitations}

RR measures local, checkpoint-specific context dependence. Layer, window, and reset prefix affect its rankings, and hidden-state cosine is not invariant to representation reparameterization. The flattened-record benchmark covers 276 English records from three domains, and its application comparison is limited to methods without task-specific model training. Exact RR also requires one reset-window evaluation per candidate edge, making it substantially more expensive than one-pass chunkers. Broader corpora, matched supervised comparators, and cheaper approximations are natural next steps.

\section{Conclusion}

Removing a prefix reveals how much the text to its right still depends on what came before. \RR{} turns that dependence into a boundary score by measuring preservation of the right-hand hidden-state trajectory, and dynamic programming turns the scores into chunks. On flattened text with weak structural cues, RR recovers substantially more source records than the tested conventional, prompted, and same-model passive baselines. The broader finding is simple: context dependence is most useful as a boundary signal where ordinary chunking cues fall short.

\section*{Competing Interests}

This research was conducted independently and does not represent the views of the author's employer. The employer had no role in the research.

\appendix

\section{Reproducibility Details}
\label{app:details}

\subsection{Models and execution}

Experiments use Python, PyTorch, and Hugging Face Transformers \citep{wolf2020transformers}. Model weights are float16 and readouts float32 on an M3 Max with 36~GB unified memory. Table~\ref{tab:revisions} lists the pinned revisions. Qwen3.5-4B is used only for the scoring-window study; Qwen3-4B-Base is used for RR in the flattened-record application. The prompted comparator uses the 4-bit MLX checkpoint \texttt{Qwen/Qwen3-4B-MLX-4bit} at revision \texttt{52a5ab34fa604bc8}.

\begin{table}[H]
\centering
\scriptsize
\caption{Pinned revisions for the six-model mechanism evaluation.}
\label{tab:revisions}
\begin{tabular}{ll}
\toprule
Model & Revision \\
\midrule
Qwen3-4B-Base & \texttt{906bfd4b4dc7f14e} \\
Qwen3.5-9B-Base & \texttt{68c46c4b3498877f} \\
Gemma 4 E4B Base & \texttt{a24c9379fd3839ae} \\
Qwen2.5-0.5B & \texttt{060db6499f32faf8} \\
GPT-2 & \texttt{607a30d783dfa663} \\
Pythia-410M & \texttt{9879c9b5f8bea905} \\
\bottomrule
\end{tabular}
\end{table}

The ten mechanism sources are four novels (\emph{Pride and Prejudice}, \emph{Moby-Dick}, \emph{Dracula}, and \emph{Dr. Jekyll and Mr. Hyde}); \emph{On the Origin of Species}; \emph{The Prince}; \emph{The Republic}; \emph{Common Sense}; the Rust ownership chapter; and a PyTorch autograd tutorial. Tokenization affects candidate packing, so each model follows the same deterministic source protocol but may not receive identical token spans. Both passages from a source remain in the same bootstrap cluster.

\subsection{Comparator definitions}

All scores are oriented so larger values prefer a cut. \method{boundary\_surprisal} is full-context NLL of the first right-hand token; \method{window\_surprisal} is mean full-context NLL over the evaluation targets; and \method{ppl\_minimum} adapts local sentence-perplexity minima \citep{zhao2024metachunking}. \method{attention\_isolation} is negative attention mass from right-window queries to prefix keys. \method{residual\_jump} is cosine distance between adjacent uninterrupted hidden states. \method{embedding\_jump} compares mean input embeddings in eight-token windows. Sentence, punctuation, fixed-grid, and seeded-random methods provide surface and policy controls.

For the application, \method{embedding\_boundary} is one minus cosine similarity between normalized BGE-base-en-v1.5 embeddings of 64-token left and right spans. \method{local\_hidden\_distance} compares mean Qwen3-4B layer-27 states in symmetric 24-token windows. Application residual jump and attention isolation use that same checkpoint and layer. The prompted comparator scores fixed batches of twelve edges using 96-token left and right contexts; 99.91\% of held-out scores are present, and the seven omissions receive the frozen fallback value. All application methods receive the same candidates, percentile transformation, penalty-calibration split, and open-count recurrence.

\subsection{Application construction}

The 60 evaluation streams use 276 unique records and the 15 calibration streams use 65 different unique records. Stream is the resampling unit within each domain; because records do not repeat, no additional repeated-record cluster is required. The primary 180-DPI condition has 99.26\% character agreement with rendered source text. Exact source joins are candidate edges in 90.7\% of cases; the constrained gold diagnostic permits nearby candidates while enforcing the common 48--384-token hard limits.

The open-count application recurrence maximizes
\[
 \sum_{b\in\C}\bigl(u(b)-\rho\bigr)
 -0.02\sum_{k}\left(\frac{t_k-t_{k-1}-192}{192}\right)^2,
\]
where the cut penalty $\rho$ is selected separately for each method using only the calibration streams. The gold record count is absent from inference.

\subsection{Conventional sectioning control}
\label{app:wiki50}

Wiki-50 is read from the original authors' archives. The 50 test documents contain 177 top-level sections and 2,776 sentences after removing the level-1 preface, folding subsections into their parent sections, and applying the original list, formula, and code-placeholder filtering. The calibration sample is the first 30 development-file members in the pinned archive order rather than a random sample; it contains 104 sections and 1,428 sentences. The two samples share no document IDs. RR uses Qwen3-4B layer 27, $W=24$, and $s=1$; BGE compares three-sentence windows on each side. Residual jump, local hidden-state distance, lexical cohesion, uniform length, and no-cut diagnostics are also recorded.

Each continuous method receives every internal sentence ending and one raw-score threshold calibrated only on the labeled development sample. No task-specific model parameters are trained. The primary metric is weighted sentence-level $P_k$ as implemented by \texttt{segeval} 2.0.11. Intervals use 10,000 document-bootstrap resamples, condition on the frozen development-calibrated thresholds, and do not include calibration-sample uncertainty.

\begin{table}[H]
\centering
\small
\caption{Wiki-50 top-level sectioning. Lower $P_k$ is better. Intervals are 95\% document-bootstrap intervals; published results do not provide comparable intervals.}
\label{tab:wiki50}
\begin{tabular}{llc}
\toprule
Method & Setting & $P_k$ [95\% CI] \\
\midrule
Human \citep{koshorek2018segmentation} & Published manual result & 0.150 \\
Supervised neural model \citep{koshorek2018segmentation} & Trained on Wiki-727K & 0.182 \\
No cuts & Diagnostic & 0.304 [0.267, 0.343] \\
\RR{} & Fixed scorer; dev threshold & 0.338 [0.275, 0.399] \\
BGE boundary & Fixed scorer; dev threshold & 0.340 [0.288, 0.377] \\
Random \citep{koshorek2018segmentation} & Published baseline & 0.527 \\
GraphSeg \citep{koshorek2018segmentation} & Published unsupervised result & 0.636 \\
\bottomrule
\end{tabular}
\end{table}

RR and BGE are statistically indistinguishable: RR minus BGE is $-0.0017$ [$-0.0467$, 0.0448]. Both undersegment, predicting 2.56 and 1.96 sections per document against a gold mean of 3.54. The no-cut diagnostic has zero boundary recall but reaches $P_k=0.304$, illustrating that sparse boundaries and false-cut penalties make severe undersegmentation competitive under this metric.

A non-operational ranking diagnostic supplies each method the gold number of sections. Under $P_k$, RR produces the best partition among the tested fixed scorers: 0.333, versus 0.363 for local hidden-state distance and 0.385 for BGE. Exact-boundary F1 is 0.157 for RR, 0.197 for local hidden-state distance, and 0.213 for BGE; mean average precision is 0.284, 0.304, and 0.301, respectively. The diagnostic therefore shows lower-$P_k$ partition geometry for RR when the count is fixed, not uniformly superior ranking of the exact gold boundaries.

\subsection{Configuration controls}
\label{app:configuration}

On Qwen2.5, output-KL separation margins at 5\%, 25\%, 50\%, 75\%, and 100\% of decoder depth are 0.120, 0.308, 0.371, 0.411, and 0.343. A Qwen3.5-4B study crosses scoring and evaluation windows of 12, 24, 36, 48, 64, and 96 tokens. All 36 combinations have positive margins; short scoring windows work best for short evaluation horizons, while long windows better predict long-horizon disruption. A 24-token score does not establish transfer to the disjoint positions 25--95: its margin is 0.0216 [$-0.0143$, 0.0614]. Replacing the empty reset with a neutral separator preserves positive matched and cross-condition margins in Qwen3, Gemma, and GPT-2. BOS and eight-token local-prefix variants produce different rankings, making the reset prefix part of the method specification.

Code, frozen configurations, source hashes, saved scores, bootstrap outputs, and figure data are available at \url{https://github.com/ZECTBynmo/right-reset-paper}. The repository README documents the released artifacts and reproduction commands.

\bibliographystyle{plainnat}
\bibliography{references}

\end{document}